\documentclass[10pt,twocolumn,letterpaper]{article}

\usepackage{cvpr}
\usepackage{times}
\usepackage{epsfig}
\usepackage{graphicx}
\usepackage{amsmath}
\usepackage{amssymb}
\usepackage{subfigure}
\usepackage{array}
\usepackage{booktabs}
\usepackage{comment}
\usepackage{multirow}

\cvprfinalcopy 

\def\cvprPaperID{7673} 

\ifcvprfinal\fi
\begin{document}

\title{Image Demoireing with Learnable Bandpass Filters}

\author{\normalsize Bolun Zheng\\
\footnotesize Hangzhou Dianzi University\\
{\tt\scriptsize zhengbolun1024@163.com}
\and
\normalsize Shanxin Yuan\\
\footnotesize Huawei Noah's Ark Lab\\
{\tt\scriptsize shanxin.yuan@huawei.com}
\and
\normalsize Gregory Slabaugh\\
\footnotesize Huawei Noah's Ark Lab\\
{\tt\scriptsize gregory.slabaugh@huawei.com}
\and
\normalsize Ale\v{s} Leonardis\\
\footnotesize Huawei Noah's Ark Lab\\
{\tt\scriptsize ales.leonardis@huawei.com}
\and
}

\maketitle

\begin{abstract}
Image demoireing is a multi-faceted image restoration task involving both texture and color restoration. 
In this paper, we propose a novel multiscale bandpass convolutional neural network (MBCNN) to address this problem. 
As an end-to-end solution, MBCNN respectively solves the two sub-problems.
For texture restoration, we propose a learnable bandpass filter (LBF) to learn the frequency prior for moire texture removal.
For color restoration, we propose a two-step tone mapping strategy, which first applies a global tone mapping to correct for a global color shift, and then performs local fine tuning of the color per pixel.
Through an ablation study, we demonstrate the effectiveness of the different components of MBCNN. 
Experimental results on two public datasets show that our method outperforms state-of-the-art methods by a large margin (more than 2dB in terms of PSNR).
\end{abstract}

\section{Introduction}
Digital screens are ubiquitous in modern daily life. We have TV screens at home, laptop/desktop screens in the office, and large LED screens in public spaces. It is becoming common practice to take pictures of these screens to quickly save information. Sometimes taking a photo is the only practical way to save information.
Unfortunately, a common side effect is that moire patterns can appear, degrading the image quality of the photo. 
Moire patterns appear when two repetitive patterns interfere with each other. In the case of taking pictures of screens, the camera’s color filter array (CFA) interferes with the screen's subpixel layout.

Unlike other image restoration problems, including denoising~\cite{zhang2017beyond}, demosaicing~\cite{DemosaicNet}, color constancy~\cite{FFCC}, sharpening~\cite{romano2016raisr}, etc., much less attention has been paid to image demoireing, which is to recover the underlying clean image from an image contaminated by moire patterns.
Only very recently, a few attempts \cite{sun2018moire, liu2018demoir, gao2019moire, he2019mop} have been made to address image demoireing. However, the problem remains to a large extent an unsolved problem, due to the large variation of moire patterns in terms of frequencies, shapes, colors, etc.

Recent works~\cite{sun2018moire, cheng2019multi, he2019mop} tried to remove moire patterns of different frequency bands through multi-scale design. 
DMCNN~\cite{sun2018moire} proposed to deal with moire patterns with a multi-scale CNN with multi-resolution branches and summed up the outputs from different scales to obtain a final output. 
MDDM~\cite{cheng2019multi} improved DMCNN by introducing an adaptive instance normalization~\cite{adain} based on a dynamic feature encoder.
DCNN~\cite{liu2018demoir} proposed a coarse-to-fine structure to remove moire patterns from two scales. The coarse scale result was upsampled and concatenated with the fine scale input for further residual learning.
MopNet~\cite{he2019mop} used a multi-scale feature aggregation sub-module to address the complex frequency, and two other sub-modules to address edges and pre-defined moire types.
Our model also adopts a multi-scale design with three branches for three different scales. Among different scales, our model adopts a gradual upsampling strategy to smoothly increase the resolution. 

\begin{figure}[t]
	\centering
	\includegraphics[width=1.0\linewidth]{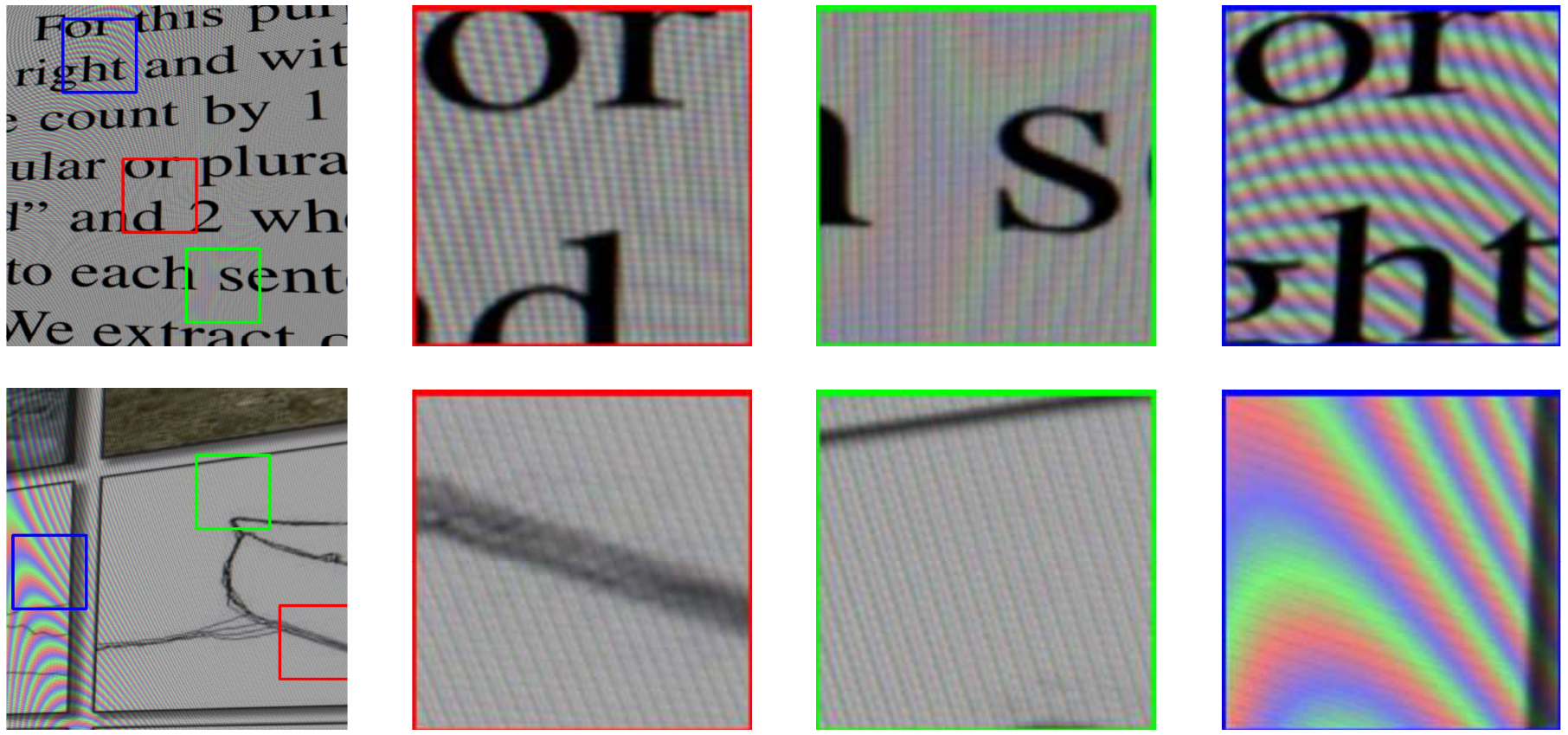}
	\caption{Moire texture of different scales, frequencies, and colors.}
	\label{MOIRE}
\end{figure}

Generally, 
none of the existing methods tried to model the moire patterns explicitly. 
In our model, we explicitly model the moire patterns by learning the frequency prior of moire patterns and respectively restore the moire image from texture and color.
Our contributions are as follows. 
\begin{itemize}
\item We introduce a unified framework namely multi-scale bandpass CNN (MBCNN) for image demoireing. The network performs both texture restoration and color restoration within the same model.
\item We propose a learnable bandpass filter (LBF) for efficient moire texture removal. The LBF introduces a learnable bandpass to learn the frequency prior, which could precisely separate moire texture from normal image texture. 
\item Our method includes global/local tone mapping for accurate color restoration. The global tone mapping learns the global color shift from moire images to clean images, while the local tone mapping is to make a local fine-grained color restoration.
\item We also propose an advanced Sobel loss (ASL) to learn the structural high-frequency information. With the ASL, we develop a multi-scale supervision to remove moire patterns in three scales.
\end{itemize}

\section{Related work}

Image demoireing requires both texture and color restoration, rendering it a complex challenge.
%
In this section, we make a brief introduction of several CNN-based methods in related tasks, where deep learning has made significant impact.

\textbf{Image restoration.} Dong \textit{et al.}~\cite{ARCNN,SRCNN} were the first to propose end-to-end convolutional neural networks for image super-resolution and compression artifact reduction. 
Subsequent research~\cite{Svoboda2016,vdsr,IRCNN} further improved these models by increasing the network depth, introducing skip connections~\cite{long2015fully} and residual learning. 
Much deeper networks~\cite{DRCN, DRRN, MemNet, RDN} were then introduced. DRCN~\cite{DRCN} proposed recursive learning for parameter sharing. Tai \textit{et al.}~\cite{DRRN,MemNet} introduced a recursive residual learning and proposed a memory block. Zhang \textit{et al.}~\cite{RDN} replaced the recursive connection in the memory block by a dense connection~\cite{DenseNet}. 
Moreover, several studies focused on multi-scale CNNs inspired by high-level computer vision methods. Mao \textit{et al.}~\cite{RED-Net} proposed a skip connection-based multi-scale autoencoder. 
Cavigelli \textit{et al.}~\cite{CAS-CNN} introduced a multi-supervised network for compression artifact reduction. 

\textbf{Frequency domain learning.} Several studies~\cite{MWCNN,DDCN,IDCN} focus on frequency domain. 
Liu \textit{et al.}~\cite{MWCNN} introduced the discrete wavelet transform and its inverse to replace conventional upscaling and downscaling operations for image restoration. 
Guo \textit{et al.}~\cite{DDCN} introduced convolution-based window sampling, Discrete Cosine Transform (DCT) and inverse DCT (IDCT) to construct a DCT-domain learning network. 
Zheng et al.~\cite{IDCN} introduced implicit DCT to extend the DCT-domain learning to color image compression artifact reduction.

\textbf{Color restoration.}  Image dehazing and image enhancement are two classic color restoration problems. 
Eilertsen et al.~\cite{HDRCNN} proposed a Gamma correction based loss function and trained a U-Net~\cite{UNet} based CNN for high dynamic range (HDR) image reconstruction. 
Gharbi \textit{et al.}~\cite{HDRNet} proposed HDRNet to learn local piece-wise linear tone mapping.  
Inspired by the guided filter~\cite{guidedfilter}, Wu \textit{et al.}~\cite{wu2018fast} proposed an end-to-end trainable guided filter for image enhancement. 
Ren et al.~\cite{ren2018gated} grouped a hazy image and several pre-enhanced images together as input, and proposed a symmetric autoencoder to learn a gated fusion for image dehazing.
Zhang et al.~\cite{zhang2018densely} proposed a densely connected pyramid CNN for image dehazing. 
Remarkably, few of these color restoration methods introduce residual connection in their solutions.

\textbf{Image demoireing.} Recently, several end-to-end image demoireing solutions have been proposed. 
Sun \textit{et al.}~\cite{sun2018moire} first introduced a CNN for image demoireing (DMCNN) and created an ImageNet~\cite{russakovsky2015imagenet}-based moire dataset for training and testing. 
Cheng \textit{et al.}~\cite{cheng2019multi} improved DMCNN by introducing an adaptive instance normalization~\cite{adain} based dynamic feature encoder.
He \textit{et al.}~\cite{he2019mop} introduced additional moire attribute labels based on shape, color, and frequency for more precise moire pattern removal.
None of the existing methods modeled the moire patterns explicitly. We treat the image demoireing problem as moire texture removal and color restoration.

\section{Proposed method}

A moire image captured by a digital camera can be modeled as:
\begin{eqnarray}
I_{moire}=\psi(I_{clean})+N_{moire}
\end{eqnarray}
where $I_{clean}$ is the clean image displayed on the screen, $N_{moire}$ is the introduced moire texture, and $\psi$ is the color degradation caused by the screen and the camera sensor. 
$I_{clean}$ can be then expressed as:
\begin{eqnarray}
I_{clean}=\psi^{-1}(I_{moire}-N_{moire})
\label{eq02}
\end{eqnarray}
where $\psi^{-1}$ is the inverse function of $\psi$, which is known as the tone mapping function in the image processing field. 
Modeled in this way, the image demoireing task can be divided into two steps, \textit{i.e.}, moire texture removal and tone mapping. 

\subsection{Multiscale bandpass CNN}

\begin{figure*}
	\centering
	\includegraphics[width=1.0\linewidth]{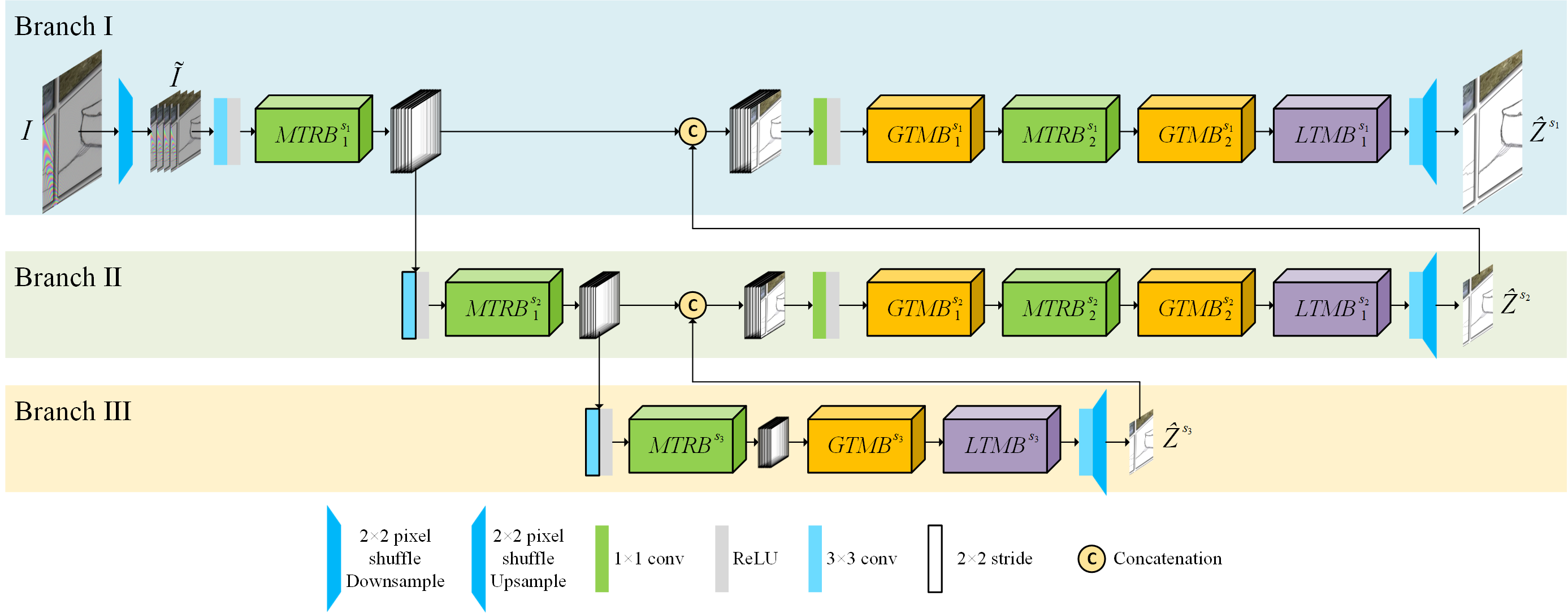}
	\caption{The architecture of our multi-scale bandpass CNN.}
	\label{MBCNN}
\end{figure*}

We propose a Multi-scale Bandpass CNN (MBCNN) to do image demoireing, \textit{i.e.}, to recover the underlying clean image from the moire image. 
Our model works in three scales and has three different types of blocks, which are moire texture removal block (MTRB), global tone mapping block (GTMB), and local tone mapping block (LTMB).  
The details of each block are described in Sec.~\ref{sec3_2} and Sec.~\ref{sec3_3}.

The architecture of MBCNN is shown in Figure~\ref{MBCNN}. 
The input image $I$ with the shape of $h\times w\times c$ is first reversibly downsampled into four subimages $\tilde{I}$ with the shape of $\frac{h}{2}\times\frac{w}{2}\times4c$. %
With the tensor $\tilde{I}$ as input, the following network consists of three branches, each to recover the moire image in a specific scale. 
Following Eq. \ref{eq02}, each branch sequentially executes the moire texture removal and tone mapping, and finally outputs an up-scaled image to be fused in the finer scale branch. 
In branch I and II, after fusing the feature of current branch and the output of the coarser scale branch, additional GTMB and MTRB are stacked to remove the texture and color errors caused by the scale change.

\subsection{Moire texture removal}
\label{sec3_2}
Moire patterns exhibit considerable variation in shape, frequency, color, \textit{etc}. Some examples are shown in Figure~\ref{MOIRE}, where the moire patterns have different characteristics. 
The moire texture can be written as:
\begin{eqnarray}
	N_{moire}=\sum_{i}\sum_{j}N_{f_{ij}}^{s_i}
	\label{eq1}
\end{eqnarray}
where $N_{f_{ij}}^{s_i}$ denotes the moire texture component of scale $s_i$ and frequency $f_{ij}$. 
Following this formulation, we can first estimate the components of moire texture at different scales and frequencies, and then reconstruct the moire texture based on all the estimated components.

Block-DCT is an effective way for handling frequency related problems.
Assuming that the frequency spectrum in block-DCT domain of each $N_{f_{ij}}^{s_i}$ is $FS_{f_{ij}}^{s_i}$, then Eq.~\ref{eq1} can be rewritten as
\begin{eqnarray}
	\begin{split}
	N_{moire}&=\sum_{i}\sum_{j}\mathcal{D}^{-1}(FS_{f_{ij}}^{s_i})\\
	&=\mathcal{D}^{-1}(\sum_{i}\sum_{j}FS_{f_{ij}}^{s_i})
	\end{split}
	\label{eq5}	
\end{eqnarray}
where $\mathcal{D}^{-1}$ denotes the block-IDCT function. 

Given a color image patch $P$, we denote the moire texture of each color channel as $N_{P}^{c}$, $c\in\{R,G,B\}$. Then the representation of the moire texture $N_{P}$ is
\begin{eqnarray}
	\mathcal{C}(N_P)=\sum_{c\in\{R,G,B\}}\mathcal{C}(N_{P}^{c})
	\label{eq2}
\end{eqnarray}
where $\mathcal{C}$ denotes a learnable convolution.
Based on Eq.~\ref{eq5}, Eq.~\ref{eq2} can be rewritten as 
\begin{eqnarray}
\begin{split}
\mathcal{C}(N_P)&=\sum_{c\in\{R,G,B\}}\mathcal{C}(\mathcal{D}^{-1}(\sum_{i}\sum_{j}FS_{f_{ij}}^{s_i}))\big|_{c}\\
&=\sum_{i}\mathcal{C}(\mathcal{D}^{-1}(\sum_{c\in\{R,G,B\}}\sum_{j}FS_{f_{ij}}^{s_i}\big|_{c}))\\
&=\sum_{i}\mathcal{C}(\mathcal{D}^{-1}(\sum_{c\in\{R,G,B\}}FS^{s_i}\big|_{c}))
\end{split}
\label{eq3}
\end{eqnarray}
where $FS^{s_i}\big|_{c}$ is the combined frequency spectrum of channel $c$ with the scale of $s_{i}$. 
Here, we define the $\sum_{c\in\{R,G,B\}}FS^{s_i}\big|_{c}$ as the implicit frequency spectrum (IFS) denoted as $\xi^{s_{i}}$. 
Now, we can have
\begin{eqnarray}
\mathcal{C}(N_P)&=\sum_{i}\mathcal{C}(\mathcal{D}^{-1}(\xi^{s_{i}}))
\label{eq4}
\end{eqnarray}

\textbf{Learnable Bandpass Filter.} 
Inspired by the implicit DCT~\cite{IDCN}, we can directly estimate  $\xi^{s_{i}}$ with a deep CNN block. 
Since the transforms presented in Eq.~\ref{eq4} are all linear, they can be modeled by a simple convolution layer. 
As the frequency spectrum of moire texture is always regular, we can use a bandpass filter to amplify certain frequencies and diminish others. 
However, it's difficult to get the frequency spectrum prior modeling the moire texture, because there would be several frequencies in different scales and they can also affect each other. 
To solve this problem, we propose a learnable bandpass filter (LBF) to learn the prior from moire images. 
LBF introduces a learnable weights for each frequency, which can be expressed as
\begin{eqnarray}
\mathcal{C}(N_P)=\sum_{i}\mathcal{C}(\mathcal{D}^{-1}(\theta^{s_{i}}\cdot\xi^{s_{i}}))
\label{eq6}
\end{eqnarray}
where $\theta^{s_{i}}$ denotes the learnable weights of DCT domain frequencies for the scale $s_{i}$. 

Assuming the size of block-IDCT is $p\times p$, then the corresponding DCT domain frequency spectrum totally has $p^2$ frequencies, so the size of $\theta^{s_{i}}$ is $p^2$. 
All parameters of  $\theta^{s_{i}}$ are initialized to be 1 and constrained to be non-negative, the passbands are learned from the image data during training. 
$\mathcal{D}^{-1}$ can be implemented by a predefined $1\times1$ convolution layer, whose weights are fixed as the IDCT matrix.

\begin{figure}[t]
	\centering
	\includegraphics[width=1.0\linewidth]{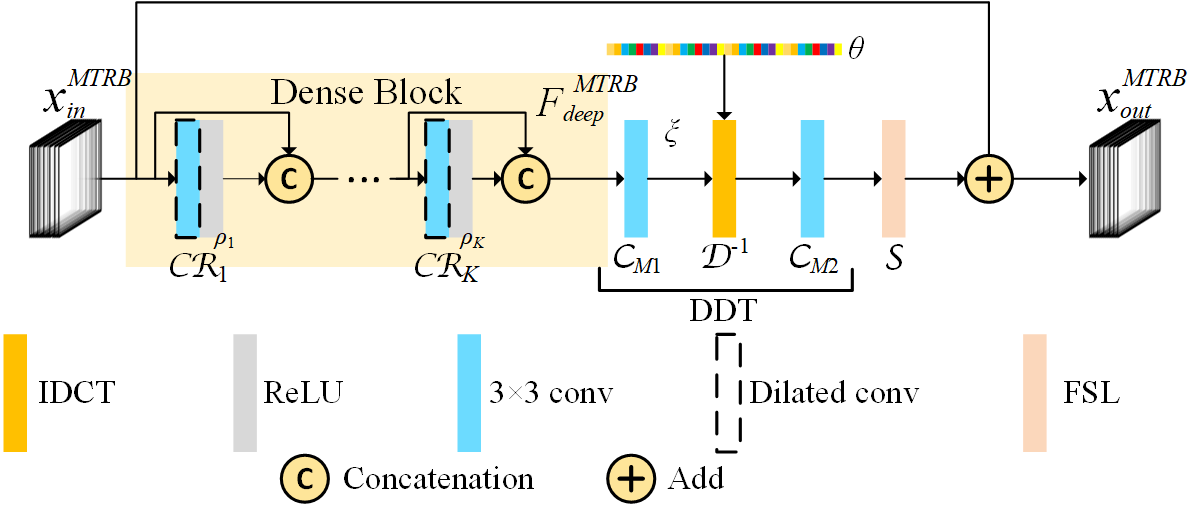}
	\caption{The structure of moire texture removal block.}
	\label{MTRB}
\end{figure}

\textbf{CNN Structure.} 
Following Eq.~\ref{eq6}, we can respectively remove moire texture 
from different scales. 
For each specific scale,  we propose a moire texture removal block (MTRB), see Figure~\ref{MTRB}. 

Assuming the input of the MTRB is $x_{in}^{MTRB}$, a dense block is first used for feature extraction, which is denoted as $F_{deep}$. Then a $3\times3$ convolution layer estimates the IFS $\xi$ from $F_{deep}$.
The dense block has $K$ densely connected~\cite{DenseNet} $3\times3$  $n_D$-channel dilated convolution~\cite{DILATED_CONV} with ReLU activation ($Conv\_ReLU$) layers. 
We adopt dilated convolution rather than normal convolution to enlarge the receptive field of the dense block to produce $F_{deep}$, so that the $p^2$ sized $\xi$ can be easily estimated from the $F_{deep}$.
After estimating $\xi$, the learnable weight $\theta$ and the block-IDCT layer $\mathcal{D}^{-1}$, a convolution layer $\mathcal{C}_{M2}$ is added as indicated in Eq.~\ref{eq6}. 

Considering that the $\mathcal{D}^{-1}$ might lead to large local output and produce excessive gradient, we stacked a Feature Scale Layer (FSL) to linearly constrain the output of $\mathcal{C}_{M2}$. 
Finally, we introduce the residual connection~\cite{ResNet} to remove the moire texture in convolution domain. 
Thus, the final output of MTRB $x_{out}^{MTRB}$ can be obtained by
\begin{eqnarray}
	x_{out}^{MTRB}=x_{in}^{MTRB}+\mathcal{S}(\mathcal{C}_{M2}(\mathcal{D}^{-1}(\theta\cdot\xi)))
\end{eqnarray}
where $\mathcal{S}$ denotes the FSL. 

Directly multiplying $\theta$ and $\xi$ will consume large amount of calculations. Instead, we reshape $\theta$ to the size of $1\times1\times p \times p$, and multiply it with the convolution kernel of $\mathcal{D}^{-1}$ layer, then the $\xi$ is directly sent to $\mathcal{D}^{-1}$ layer. In this way, the product $\theta\cdot\xi$ can be avoided.

\subsection{Tone mapping}
\label{sec3_3}
The RGB color space is an extremely large space containing $256^3$ colors, making it difficult to do point-wise tone mapping. 
Observing that there are color shifts between the moire and clean images, we  propose a two-step tone mapping strategy with two types of tone mapping blocks: Global Tone Mapping Block (GTMB) and Local Tone Mapping Block (LTMB).

\begin{figure}[t]
	\centering
	\includegraphics[width=1.0\linewidth]{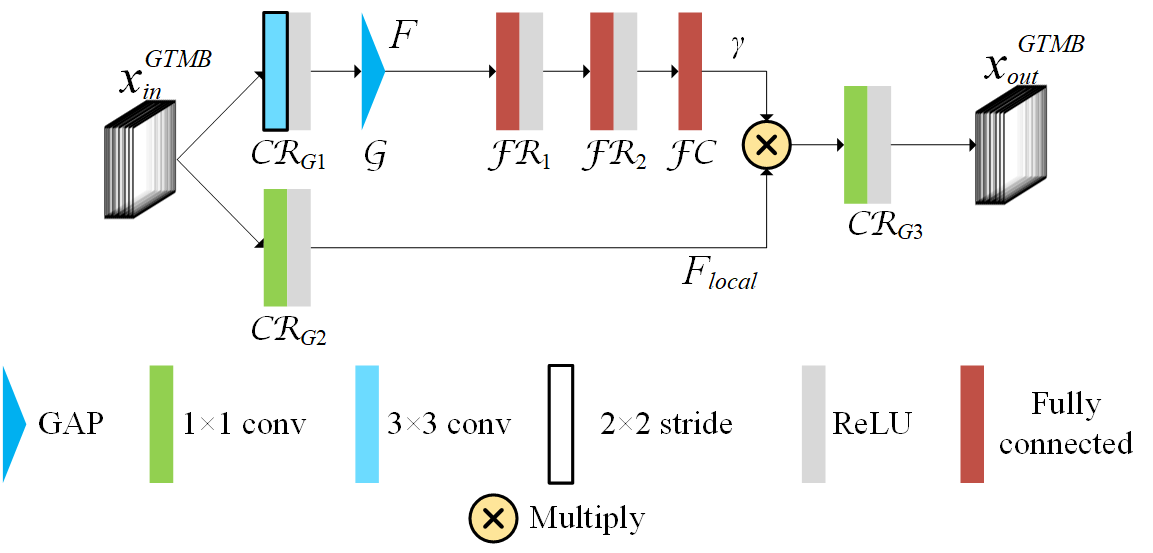}
	\caption{The structure of global tone mapping block.}
	\label{GTMB}
\end{figure}

\begin{table}[ht]
\normalsize 
  \centering
  \resizebox{1.0\columnwidth}{!}{
  \begin{tabular}{lcccccc}
  \toprule 
  Layer &
  $\mathcal{CR}_{G1}$ & 	$\mathcal{CR}_{G2}$ &
  $\mathcal{CR}_{G3}$ &
  $\mathcal{FR}_{1}$  &
  $\mathcal{FR}_{2}$  &
  $\mathcal{FC}$ \\
	\midrule
	Stride       & $2\times2$ & $1\times1$ & $1\times1$ & - & - & -\\
	Kernel     & $3\times3$ & $1\times1$ & $1\times1$ & - & - & -\\
	Output Ch. & $n_{G}\cdot2$ & $n_{G}\cdot2$ & $n_{G}$ & $n_{G}\cdot8$ & $n_{G}\cdot4$ & $n_{G}\cdot2$\\
  \bottomrule
  \end{tabular}}
	\caption{Attributions of learnable layers in GTMB.}
	\label{table1}
\end{table}
\textbf{Global tone mapping block.} 
The GTMB is proposed to learn the global color shift, see Figure~\ref{GTMB} for the detailed structure. 
Given the input $x_{in}^{GTMB}$, we first extract a global feature $F$ through a $3\times3$ $Conv\_ReLU$ layer with the stride of 2 and a global average pooling (GAP) layer.
Then, to extract a deep global feature $\gamma$, we stack two fully connected (FC) layers with ReLU activation ($\mathcal{FR}_{1}$, $\mathcal{FR}_{2}$) and a FC layer without ReLU activation ($\mathcal{FC}$).
Besides, we use an $1\times1$ $Conv\_ReLU$ layer extracts the local feature $F_{local}$ from $x_{in}^{GTMB}$.
The output of GTMB $x_{out}^{GTMB}$can be obtained as
\begin{eqnarray}
x_{out}^{GTMB} = \mathcal{CR}_{G3}(\gamma\cdot F_{local})
\end{eqnarray}
Assuming the $\mathcal{CR}_{G3}$ outputs a $n_{G}$-channel tensor, Table~\ref{table1} lists the attributions of all learnable layers in GTMB.

\textbf{GTMB vs. Channel Attention.} The attention mechanism has proven to be effective in many tasks\cite{yan2019stat,wang2019edvr,yan2020TPMAI, 3droom}, and several channel attention blocks have been proposed \cite{rcan, hu2018squeeze}. Our GTMB can be view as a channel attention block.
However, GTMB is different from existing channel attention blocks in several aspects. 
First, existing channel attention blocks are always activated by a Sigmoid unit, while there are no such constraints for the $\gamma$ in GTMB. 
Second, channel attention is directly applied on the input of the existing channel attention blocks,  while the $\gamma$ in GTMB is applied on the local feature $F_{local}$.
Finally, existing channel attention blocks are aimed at making an adaptive channel-wise feature re-calibration; the goal of GTMB is to make a global color shift and avoid the irregular and inhomogeneous local color artifacts (more analysis are described in Sec. \ref{sec4_4_1}).
%

\begin{figure}[t]
	\centering
	\includegraphics[width=1.0\linewidth]{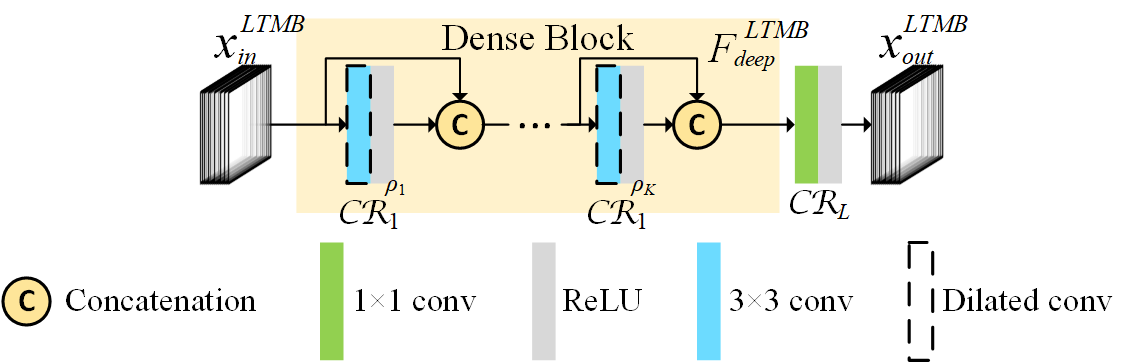}
	\caption{The structure of local tone mapping block.}
	\label{LTMB}
\end{figure}
\textbf{Local tone mapping block.} 
The LTMB is developed to fit a local fine-grained tone mapping function. 
As shown in Figure~\ref{LTMB}, the structure of LTMB is similar to MTRB. 
LTMB first takes a similar dense block in MTRB to extract the deep feature $F_{deep}^{LTMB}$ from the input of LTMB $x_{in}^{LTMB}$. 
Then, the output of LTMB is obtained by 
\begin{eqnarray}
	x_{out}^{LTMB}=\mathcal{CR}_{L}(F_{deep}^{LTMB})
\end{eqnarray}
where $\mathcal{CR}_{L}$ is a $1\times1$ convolution, and $x_{out}^{LTMB}$ has the same shape with $x_{in}^{LTMB}$.

\subsection{Loss function}

\begin{figure}
	\centering
	\subfigure[]{
		\begin{minipage}[htbp]{0.45\linewidth}
			\includegraphics[width=\textwidth]{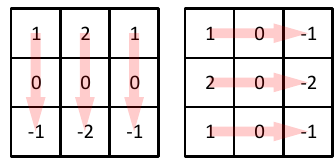}	
		\end{minipage}
		\label{ASL-a}
	}
	\subfigure[]{
		\begin{minipage}[htbp]{0.45\linewidth}
			\includegraphics[width=\textwidth]{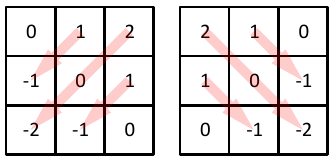}	
		\end{minipage}
		\label{ASL-b}
	}
	\caption{Details of advanced Sobel loss. (a) Classic Sobel filters. (b) Two additional filters for advanced Sobel filters.}
	\label{ASL}
\end{figure}

In this paper, we use the L1 loss as the base loss function, as it has been proven \cite{lim2017enhanced,RDN,zhao2016loss} that L1 loss is more effective than L2 loss for image restoration tasks.
However, the L1 loss itself is not enough as it is a point-wise loss that cannot provide structural information, while moire patterns are structural artifact. 
We propose an Advanced Sobel Loss (ASL) to solve this problem. 
The proposed ASL can be expressed as
\begin{eqnarray}
\mathcal{ASL}(\hat{Z}, Z)=\dfrac{1}{N}\sum\big|Sobel^{*}(Z)-Sobel^{*}(\hat{Z})\big|
\end{eqnarray}
where $Z$ denotes the groundtruth, $\hat{Z}$ denotes the output of CNN, and $Sobel^*$ denotes the advanced Sobel filtering. Figure~\ref{ASL} illustrates the details of ASL.
Compared to classic Sobel filters (Figure~\ref{ASL-a}), the advanced Sobel filters provide two additional  filters of $45^\circ$ directions (Figure~\ref{ASL-b}), which could provide richer structure information. 
We combine ASL and L1 loss as the final loss function, which can be expressed as,
\begin{eqnarray}
\label{eq13}
Loss(\hat{Z}, Z)=\mathcal{L}1(\hat{Z}, Z)+\lambda\cdot\mathcal{ASL}(\hat{Z}, Z)
\end{eqnarray}
where $\mathcal{L}1$ denotes the L1 loss, $\mathcal{ASL}$ denotes the ASL, and $\lambda$ is a hyper-parameter to balance the L1 loss and ASL.
 
When training MBCNN, we adopt the multi-supervising strategy that supervising the outputs from all branches, which can be expressed as, 
\begin{eqnarray}
\label{eq14}
\begin{split}
	loss = Loss(\hat{Z}^{s_1}, Z^{s_1})&+Loss(\hat{Z}^{s_2}, Z^{s_2})\\
	&+Loss(\hat{Z}^{s_3}, Z^{s_3})
\end{split}
\end{eqnarray}
where $s_{1}$, $s_{2}$, and $s_{3}$ indicate branch 1, 2, and 3, respectively. 

\section{Experiments} 
\label{EXP_SEC}

We have conducted extensive ablation studies and outperformed state-of-the-art by large margins on two public datasets: \emph{LCDMoire}~\cite{AIM19demoireDataset} and \emph{TIP2018}~\cite{sun2018moire}
The \emph{LCDMoie} dataset consists of 10,200 synthetically generated image pairs with 10,000 training images, 100 validation images and 100 testing images.
The \emph{TIP2018} dataset consists of real photographs constructed by photographing images of the ImageNet~\cite{russakovsky2015imagenet} dataset displayed on computer screens with various combinations of different camera and screen hardware. It has 150,000 real clean and moire image pairs, split into 135,000 training images and 15,000 testing images.
Both \emph{LCDMoire} and \emph{TIP2018} datasets are used to do comparison with state-of-the-art methods. \emph{LCDMoire} dataset is also used for ablation study. The ablation study is conducted on the validation set, as the test dataset's ground truth is not available. Please note: the validation dataset is completely independent and not used in training.

\subsection{Implementation details}
\label{imp details}
For the MBCNN model, we adopt the following settings, with $c=3$, $n_G=128$, $n_D=64$, $K=5$. 
Adam~\cite{adam} is used as our training optimizer. The learning rate is initialized to be $10^{-4}$. The validation was conducted after every training epoch. If the decrease in the validation loss was lower than 0.001 dB for four consecutive epochs, the learning rate was halved. 
When the learning rate became lower than $10^{-6}$, the training procedure was completed. 
For \emph{LCDMoire} dataset, we $128\times128$ patches were randomly cropped from the images, with the batch size set to 16. When the $128\times128$ patch trained model converged, we re-grouped the training data into $256\times256$ patches for fine-tuning the model. This time, the learning rate was set to $10^{-5}$, the batch size was set to 4. Training a MBCNN roughly takes 40 hours with a NVidia RTX2080Ti GPU.
For \emph{TIP2018} dataset, we follow~\cite{sun2018moire} and set the patch size as $256\times256$ through out the training. 

\subsection{Ablation Study}

To verify the effectiveness of each component in our model, we conduct extensive ablation studies, including evaluation of MTRB vs. GTMB and LTMB, learnable bandpass filter, and loss function.

\begin{figure}[t]
  \centering
  \includegraphics[width=1.0\linewidth]{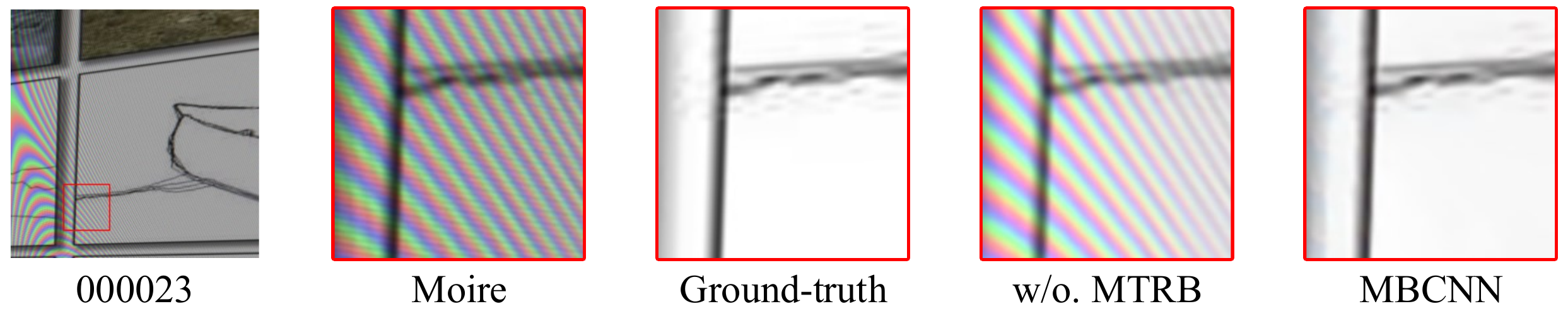}
  \caption{Demoireing results produced by MBCNN with and without MTRB.}
  \label{MTRB-Comp}
\end{figure}

\subsubsection{MTRB vs. GTMB and LTMB}

As described in previous sections, the 
MTRB is designed for removing moire texture, GTMB and LTMB are designed for color restoration. 
We investigate the effect of the MTRB using a trained MBCNN, and visualize the experimental results in Figure.~\ref{MTRB-Comp}. 
Due to the residual connection in MTRB, we can separate the effect of MTRB from the two tone mapping blocks by forcing the learned scale in the feature scaling layer to be zero. 
As shown in Figure~\ref{MTRB-Comp}, without MTRBs, the degraded color can still be well restored, and some of very high frequency moire texture can also be well removed. 
However many high frequency image details are lost, and the low-frequency moire texture largely remains. 
The result is mainly caused by two reasons. 
First, because $3\times3$ convolutions are used in GTMB and LTMB, the CNN has certain denoising and local smoothing capabilities. 
Second, although the proposed tone mapping blocks do have a great ability to restore color, the major contribution to moire texture removal is made by MTRBs. 
This experiment demonstrates that the MTRBs have strong capability to do moire texture removing, while the GTMBs and LTMBs are good at restoring colors. 


\subsubsection{Learnable bandpass filter}
In this section, we investigate the contribution of LBF and explain the reasons why we choose the relevant settings. 

\begin{table}[h]
\small
  \centering
  \resizebox{1.0\columnwidth}{!}{
  \begin{tabular}{lccc}
  \toprule 
  	Model & MBCNN-nDDT & MBCNN-nLP & MBCNN\\
  \midrule 
			PSNR/SSIM & 42.91/0.9932 & 43.09/0.9936 & 44.04/0.9948\\
  \bottomrule
  \end{tabular}}
	\caption{Performance of MBCNN, MBCNN-nLP and MBCNN-nDDT on \textit{LCDMoire} validation set.}
	\label{table2}
\end{table}
\textbf{Structural contribution.} 
The LBF is constructed by two parts, DCT domain transform (DDT) and the learnable passband (LP). 
We applied the settings described in Section~\ref{imp details}, and respectively removed the DDT and LP from the MTRBs to conduct the investigation. 
We removed the entire DDT by replacing it by a $1\times1$ convolution layer to keep the output shape unchanged. In this case, the MTRB degenerates to a residual dense block (RDB). 
We removed the LP by keeping the entire DDT, but forcing all parameters in the passbands to be 1, which will not be updated during training phase.

\begin{figure}[t]
	\centering
	\includegraphics[width=1.0\linewidth]{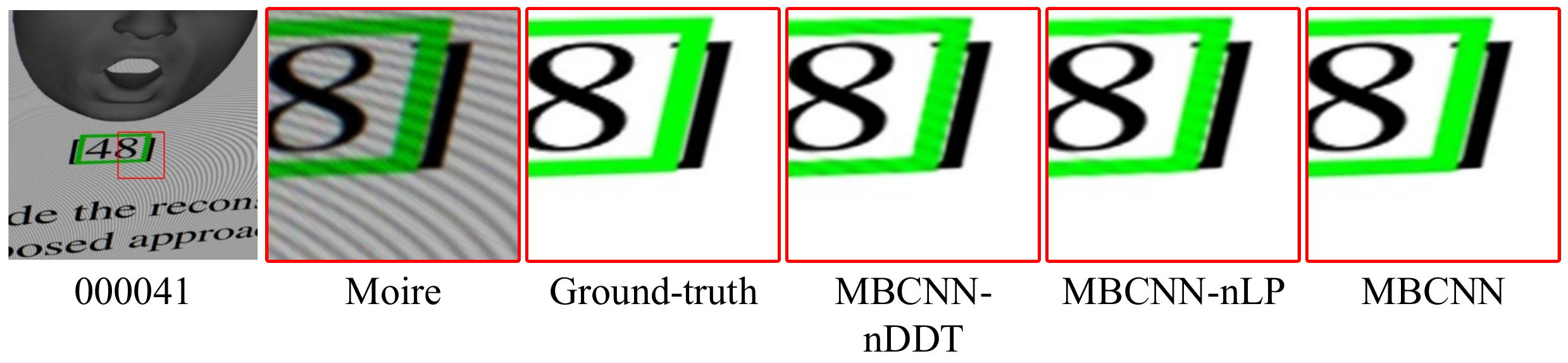}
	\caption{Demoireing results produced by MBCNN-nDDT, MBCNN-nLP and MBCNN.}
	\label{demoire_comp_1}
\end{figure}

We denote the networks constructed without LP or DDT as MBCNN-nLP and MBCNN-nDDT, respectively. 
We tested the performance of these three models on the validation set of \emph{LCDMoire}. 
As shown in Table \ref{table2}, MBCNN-nLP introduces the DDT which could provide a structural learning path and explicitly ensure the internal receptive field (block-IDCT size), and finally leads to a slight improvement of 0.18dB from MBCNN-nDDT. 
MBCNN introduces the learnable bandpass to learn the frequency prior of the moire texture and leads a significant improvement of 0.95 dB from MBCNN-nLP. 
Some demoireing results produced by these three models are shown in Figure~\ref{demoire_comp_1}. 
The LBFs enable the MBCNN to better sense the moire texture and recover more accurate details from moire images.

\begin{table}[ht]
\normalsize 
  \centering
  \resizebox{1.0\columnwidth}{!}{
  \begin{tabular}{lcccc}
  \toprule 
   Model & MBCNN-6 & MBCNN-8 & MBCNN-10 & MBCNN-12\\
  \midrule 
   PSNR/SSIM & 43.25/0.9937 & 44.04/0.9948 & 43.45/0.9939 & 43.17/0.9937 \\
  \bottomrule
  \end{tabular}}
  \caption{Comparison of MBCNNs with different $p$ values.}
	\label{table3}
\end{table}

\begin{figure}[t]
	\centering
	\includegraphics[width=1.0\linewidth]{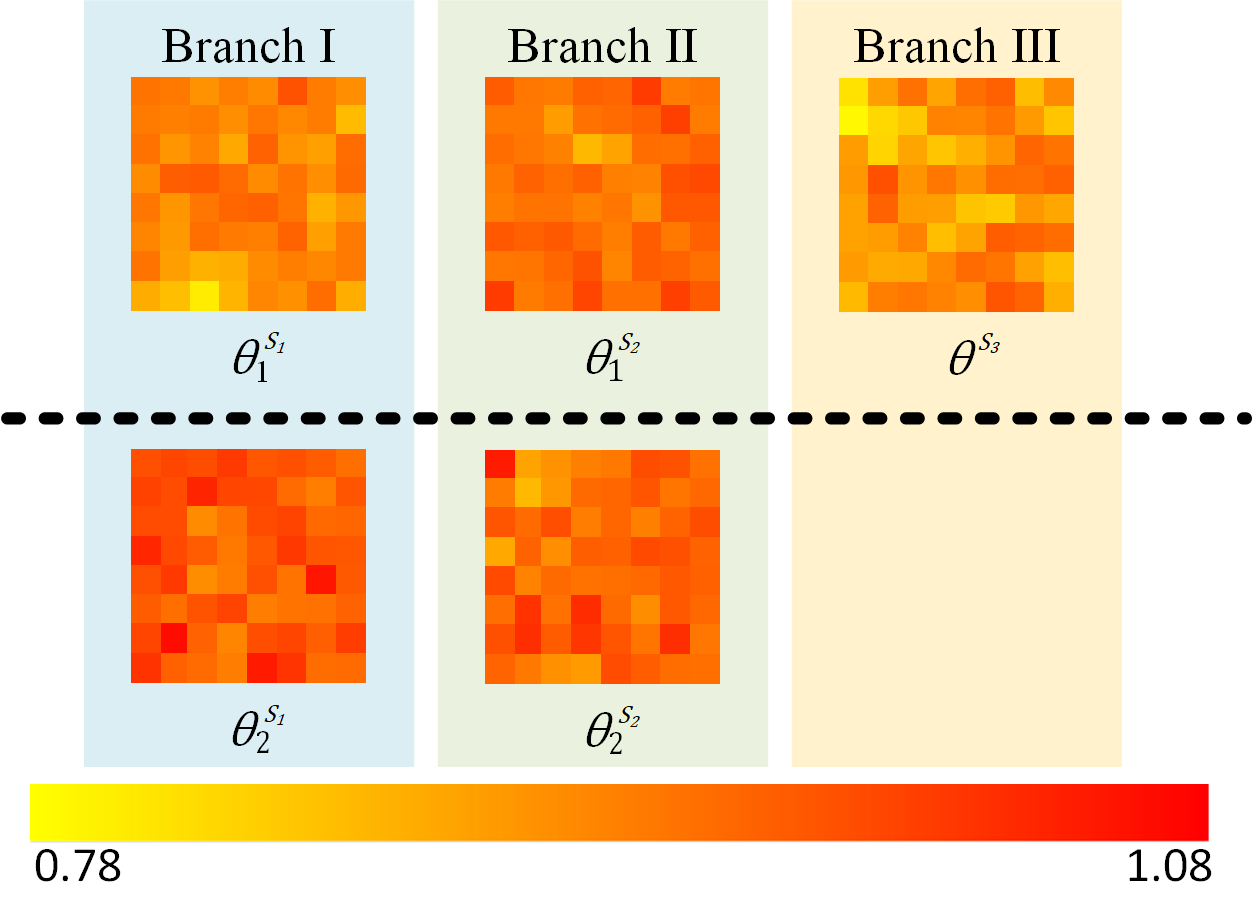}
	\caption{The learned frequency domain priors from the LBFs in different MTRBs.}
	\label{passbands}
\end{figure}

\textbf{Block-IDCT size $p$.}
$p$ is a very important parameter for DDT. With a larger $p$, the LBF can learn a more accurate and more complete frequency prior. 
We denoted the MBCNN constructed with the block-IDCT size of $p$ as MBCNN-$p$. 
We respectively validated the performance of MBCNNs constructed with $p=6, 8, 10, 12$. $p=8$ is found to be the best for moire texture removal. 
As shown in Table~\ref{table3}, larger $p$ doesn't always lead to a better result. 
There are two reasons for this observation. First, enlarging $p$ increases the complexity and difficulty of the frequency prior learning. 
Second, the receptive field provided by the front dense block cannot support a $p$ that is too large. 
We visualize the learned passbands in the LBFs from an MBCNN-8 model in Figure~\ref{passbands}. 
The LBFs perform band suppression mainly at the beginning of the branches.
The LBFs at the end of the branches are primarily avoiding over-smoothing caused by concatenating the output from the upper scale.

\subsubsection{Study of the loss function}
\label{loss_study}
In this subsection, we investigate the contribution from the loss functions. To demonstrate the effectiveness of the proposed ASL, we compare it with several related and well-known loss functions, including Sobel loss, Laplace loss, SSIM loss~\cite{zhao2016loss} and perceptual loss basing on pre-trained Vgg16 network~\cite{vggloss}.
Generally, all loss function are loaded through the multi-supervising strategy stated in Eq. \ref{eq14} and finally measured by an MAE function. 
To balance the outputs of these losses and L1 loss, we assigned different $\lambda$ (in Eq, \ref{eq13}) to different losses.
As shown in Table~\ref{table4}, the structural high frequency loss provided by the Sobel loss leads to a significant improvement of 1.81dB, and the additional two directional filters from ASL further improve the performance of 0.40dB. 
Though Laplace loss is also a high frequency descriptor, because it has a much higher weight on the center pixel than the neighbouring pixels, it behaves similar to the L1 loss.
Besides, the SSIM loss and perceptual loss also can improve the performance. The SSIM loss behaves similar to Laplace loss, while the perceptual loss is the second best loss function which is only 0.21 dB inferior to ASL. 
Generally, our ASL is an simple and effective loss function for image demoireing task.

\begin{table}[ht]
\normalsize 
  \centering
  \resizebox{0.9\columnwidth}{!}{
  \begin{tabular}{lccc}
  \toprule 
   	Loss & $\lambda$ & PSNR (dB) & SSIM\\
  \midrule 
   L1  & - & 41.83 & 0.9905\\
   L1 + Sobel & 0.5 & 43.64 & 0.9945 \\
   L1 + Laplace & 0.5 & 42.92 & 0.9927 \\
   L1 + SSIM & 0.2 & 43.36 & 0.9946 \\
   L1 + perceptual & 1.0 & 43.83 & 0.9946 \\
   L1 + ASL & 0.25 & 44.04 & 0.9948 \\
  \bottomrule
  \end{tabular}}
    		\caption{Performance comparison of MBCNN models trained with different loss functions.}
	\label{table4}
\end{table}

\begin{figure}
	\centering
	\includegraphics[width=1.0\linewidth]{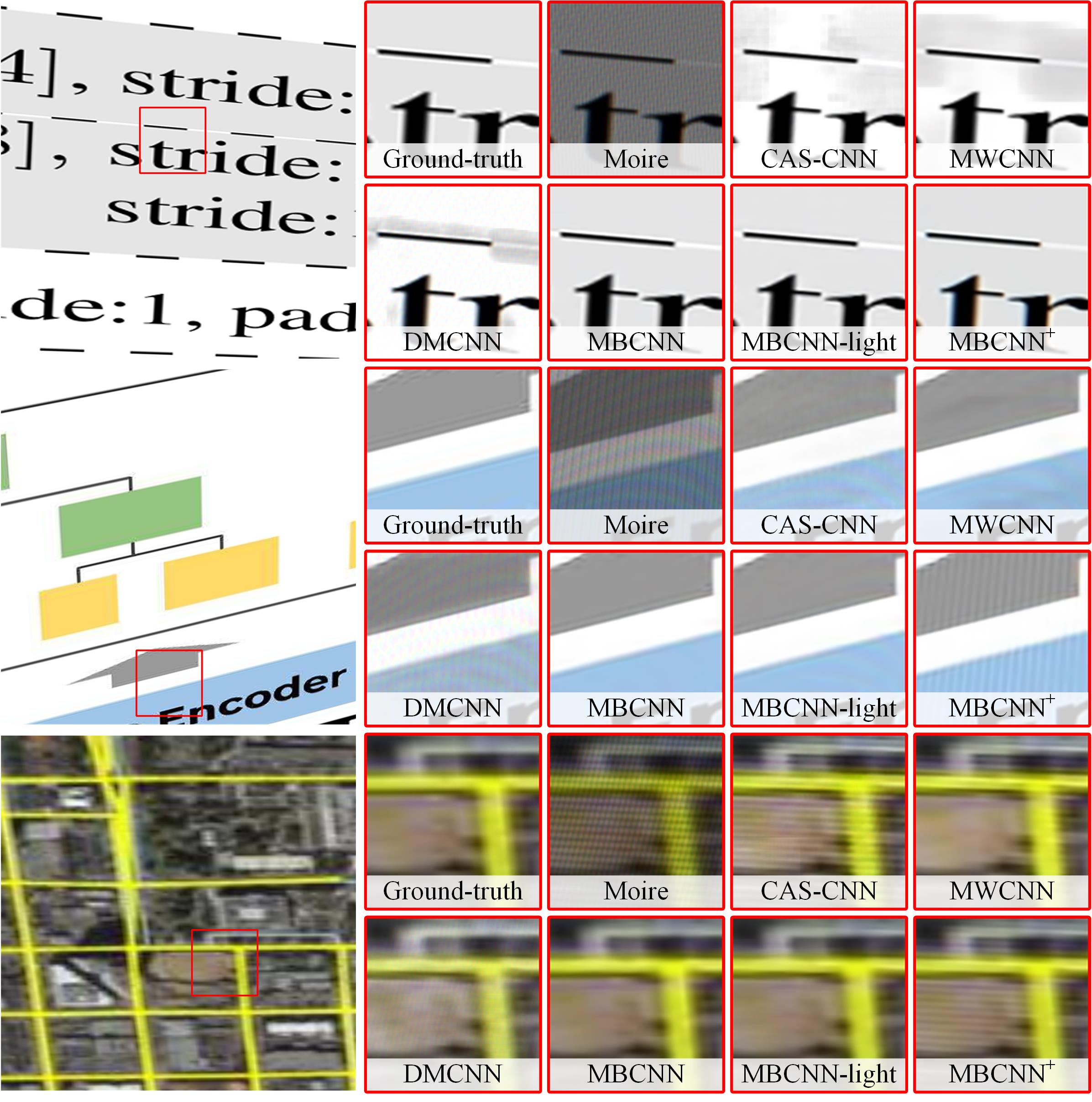}
	\caption{Demoireing results on the validation set of \emph{LCDMoire} produced by proposed methods and other prior mehods.}
	\label{LCDM}
\end{figure}

\begin{figure*}[t]
	\centering
	\includegraphics[width=1.0\linewidth]{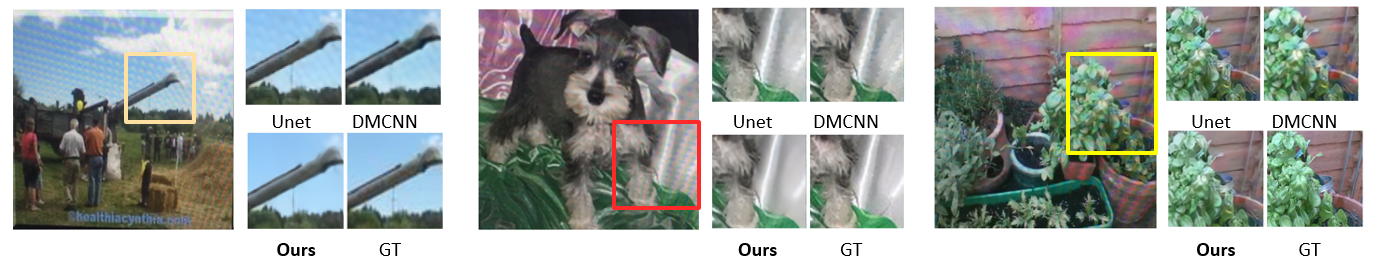}
	\caption{Qualitative comparison on \textit{TIP2018} dataset.}
	\label{pic:TIP}
\end{figure*}

\subsection{Comparison with prior work}
\label{comp_prior}
In this subsection, we compare the proposed method with several most related prior work. 

\begin{table*}[t]
\normalsize 
  \centering
  \resizebox{2.\columnwidth}{!}{
  \begin{tabular}{lccccccccc}
  \toprule 
   	Model & IPCV IITM & PCALab & IAIR & XMU-VIPLab & KU-CVIP & MoePhoto & Islab-zju & MBCNN\\
  \midrule 
	
		PSNR/SSIM  &
		32.23/0.96 & 
		32,39.0.97 & 
		35.27/0.97 &
		39.21/0.99 &
		40.17/0.98 &
		41.91/0.99 &
		42.90/0.99 &
	    \textbf{44.04/0.9948}\\
  \bottomrule
  \end{tabular}}
 	\caption{Performance comparison of MBCNN models and the top 7 participating methods in the AIM19 demoireing challenge.}
	\label{tab:challenge}
\end{table*}

\begin{table}[th]
\normalsize 
  \centering
  \resizebox{1.\columnwidth}{!}{
  \begin{tabular}{lccccccc}
  \toprule 
   	Model & CAS-CNN & MWCNN & DMCNN & MBCNN & MBCNN-light & $\text{MBCNN}^{+}$\\
  \midrule 
	PSNR &  36.16 & 28.93 & 35.48 & 44.04 & 42.81 & 33.65\\
	SSIM &  0.9873 & 0.9698 & 0.9785 & 0.9948 & 0.9940 & 0.9859\\
	Time(s)  & 0.14 & 0.14 & 0.10 & 0.25 & 0.12 & 1.14\\
  \bottomrule
  \end{tabular}}
 	\caption{Performance comparison of MBCNN models and other prior work on the validation set of \emph{LCDMoire}.}
	\label{table5}
\end{table}

\begin{table}[h]
\normalsize 
  \centering
  \resizebox{1.\columnwidth}{!}{
  \begin{tabular}{lcccccccc}
  \toprule 
   & DnCNN & VDSR & EDSR & UNet & DMCNN & MopNet & \textbf{MBCNN}  \\
  \midrule 
PSNR  & 24.54 & 24.68 &  26.82
 & 26.49 & 26.77 & 27.75 & \textbf{30.03} \\
SSIM & 0.834 & 0.837 &  0.853  & 0.864 & 0.871 & \textbf{0.895} & 0.893  \\
  \bottomrule
  \end{tabular}}
    	\caption{Performance comparison of MBCNN models and other related works on \emph{TIP2018} dataset.}
	\label{table6}
\end{table}

\subsubsection{Comparison on LCDMoire dataset}
\label{sec4_4_1}
We first compare with the participating methods in the AIM19 image demoireing challenge~\cite{AIM19demoireMethods}. The results on the validation set (again, independent and not used in training) is shown in Table~\ref{tab:challenge}.
Since the ground-truth of the \emph{LCDMoire} testing set is not released, we provide the performance on the \emph{LCDMoire} validation set.   
We also compared with several methods that did not participate in the challenge, including CAS-CNN~\cite{CAS-CNN}, MWCNN~\cite{MWCNN}, DMCNN~\cite{sun2018moire}. The result and average running time per image are shown in Table~\ref{table5}.  
Because we have demonstrated the superiority of the ASL, we trained the methods (CAS-CNN, MWCNN, DMCNN) with L1 loss plus ASL. 
Limited by the global residual connection, MWCNN fails to solve the image demoireing problem, while CAS-CNN achieves a very close performance to DMCNN. 
The proposed MBCNN method clearly outperforms these other methods, with a significant performance gain of $+7.88$dB/$+0.075$ PSNR than CAS-CNN. 
From the visualized results shown in Figure~\ref{LCDM}, our MBCNN accurately removes moire texture and restores most image details. 

However, since MBCNN consumes considerable parameters compared to several compared methods, we propose a light version of MBCNN (MBCNN-light) by setting $n_{G}=64$, $n_{D}=32$, while keeping other settings unchanged. 
As shown in Table~\ref{table5}, the fewer parameters leads to a performance reduction of $-1.46$ dB/$-0.028$ from MBCNN. 
Nevertheless, MBCNN-light still outperforms other participating methods even in this reduced form of the method.

Recently, several studies have reported that the geometric self-ensemble could reasonably enhance the performance in the final testing phase. 
We adopted this strategy during testing time by rotating the input image by $90^\circ$, $180^\circ$ and $270^\circ$ to generate three augmented input images, and calculating the mean image of the original output and three augmented outputs (rotated back) as the final output. 
We denoted this self-ensemble MBCNN as $\text{MBCNN}^+$. 
Perhaps surprisingly, this strategy leads to a dramatic reduction in performance. 
We speculate that because the moire texture is a strongly direction-aware artifact, changing the direction would mislead the network to make an inaccurate restoration.

\subsubsection{Comparison on TIP2018 dataset}
Since some related work is evaluated on the \emph{TIP2018} dataset, we further evaluated our MBCNN on the \emph{TIP2018} dataset to compare with several related methods including
DnCNN \cite{zhang2017beyond}, VDSR \cite{kim2016accurate}, EDSR \cite{lim2017enhanced}, UNet \cite{UNet}, DMCNN \cite{sun2018moire}, MopNet \cite{he2019mop}.
As shown in Table~\ref{table6}, our proposed MBCNN beats the second best method by $+2.28$ dB, in terms of PSNR, and achieved the second best SSIM result which is only 0.002 lower than the best. 
Moreover, the visualized results shown in Figure~\ref{pic:TIP} also demonstrates the proposed method outperformed other compared methods. More qualitative examples are shown in the supplementary material.

\section{Conclusion}

In this paper, we propose a multiscale bandpass CNN (MBCNN) for image demoireing, and significantly outperform state-of-the-art methods by more than 2dB in terms of PSNR. A learnable bandpass filter (LBF) is proposed to learn the frequency prior. Our model has two steps: moire texture removal and tone mapping. A LBF-based residual CNN block is used for moire texture removal, and another two CNN blocks for global and local tone mappings. An ablation study was conducted to show the importance of the components in the network. We have also clarified the the effect of the block-IDCT size in the LBF, and demonstrated that the block-IDCT size of 8 is the best for the image demoireing task. Experiments on two public datasets show that our model outperformed state-of-the-art methods by large margins.


{\small
\bibliographystyle{ieee_fullname}
\bibliography{egbib}
}

\end{document}